\documentclass[10pt,twocolumn,letterpaper]{article}

\usepackage{xcolor}
\usepackage{cvpr} 
\definecolor{iccvblue}{rgb}{0.21,0.49,0.74}
\definecolor{iccvred}{rgb}{0.8,0.0,0.0} 
\usepackage[pagebackref,breaklinks,colorlinks]{hyperref}
\hypersetup{
    linkcolor=blue,  
    citecolor=blue,  
    filecolor=iccvblue,  
    urlcolor=iccvblue,   
    allcolors=blue   
}

\AtBeginDocument{
    \hypersetup{
        linkcolor=red 
    }
}

\usepackage{pifont}
\usepackage{booktabs}
\usepackage{array}
\usepackage{graphicx}
\usepackage{graphicx}
\usepackage{amsmath}
\usepackage{amssymb}
\usepackage{booktabs}
\usepackage{amsfonts,amssymb}
\usepackage{algorithm}
\usepackage{algpseudocode}
\usepackage{multirow}
\usepackage{float}
\usepackage{caption}
\usepackage{subcaption}
\usepackage{fontawesome}
\usepackage{tabularx,caption}
\usepackage[rightcaption]{sidecap}
\usepackage{colortbl}
\usepackage{psfrag}
\usepackage[percent]{overpic}
\usepackage{url}
\usepackage[accsupp]{axessibility}
\usepackage{tablefootnote}
\usepackage{caption} %
\newcommand{\tabincell}[2]{\begin{tabular}{@{}#1@{}}#2\end{tabular}}
\newcommand{\red}[1]{\textcolor{red}{#1}}
\newcommand{\black}[1]{\textcolor{black}{#1}}
\newcommand{\blue}[1]{\textcolor{blue}{#1}}
\def\ie{\textit{i.e., }}
\definecolor{mygray}{gray}{.9}
\definecolor{lightblue}{RGB}{220, 230, 255} 

\usepackage{threeparttable}

\usepackage{tikz} 

\newcommand{\bigcircle}[1]{\tikz[baseline=(char.base)]{
    \node[shape=circle, draw, fill=black!75, text=white, inner sep=0.4pt, minimum size=4pt] (char) {\textbf{#1}};}}

\def\httilde{\mbox{\tt\raisebox{-.5ex}{\symbol{126}}}}

\begin{document}

\title{VOIDFace: A Privacy-Preserving Multi-Network Face Recognition With Enhanced Security}

\author{Ajnas Muhammed\textsuperscript{1}, Iurii Medvedev \textsuperscript{1}, Nuno Gonçalves\textsuperscript{1} \\
\textsuperscript{1}{Institute of Systems and Robotics, University of Coimbra, Coimbra, Portugal, 3030-290}\\
{\tt\small \{ajnas.muhammed, iurii.medvedev\}@isr.uc.pt, nunogon@deec.uc.pt}\\
}

\twocolumn[{%
\renewcommand\twocolumn[1][]{#1}%
\maketitle%
\vspace{-2em}%
}]
\maketitle
\thispagestyle{empty}

\begin{abstract}
 Advancement of machine learning techniques, combined with the availability of large-scale datasets, has significantly improved the accuracy and efficiency of facial recognition. Modern facial recognition systems are trained using large face datasets collected from diverse individuals or public repositories. However, for training, these datasets are often replicated and stored in multiple workstations, resulting in data replication, which complicates database management and oversight. Currently, once a user submits their face for dataset preparation, they lose control over how their data is used, raising significant privacy and ethical concerns. This paper introduces VOIDFace, a novel framework for facial recognition systems that addresses two major issues. First, it eliminates the need of data replication and improves data control to securely store training face data by using visual secret sharing. Second, it proposes a patch-based multi-training network that uses this novel training data storage mechanism to develop a robust, privacy-preserving facial recognition system. By integrating these advancements, VOIDFace aims to improve the privacy, security, and efficiency of facial recognition training, while ensuring greater control over sensitive personal face data. VOIDFace also enables users to exercise their \textit{Right-To-Be-Forgotten} property to control their personal data. Experimental evaluations on the VGGFace2 dataset show that VOIDFace provides \textit{Right-To-Be-Forgotten}, improved data control, security, and privacy while maintaining competitive facial recognition performance.
 Code is available at: \url{https://github.com/ajnasmuhammed89/VOIDFace}
\end{abstract}

\section{Introduction} \label{sec:intro}

Facial Recognition (FR) uses facial patterns to automatically identify or verify an individual. In the era of Artificial Intelligence (AI), rapid advancement of various learning models, hardware, and large dataset availability significantly enhances FR accuracy and performance. Due to this, FR has become an integral part of a wide range of security applications, ranging from smartphone unlocking \cite{opanasenko2024ensemble} to immigration and border checks \cite{hidayat2024face}. 

With widespread popularity, a significant concern coming out is the protection of highly sensitive face data. Different regulatory bodies such as the AI act \cite{edwards2021eu} and EU General Data Protection Regulation (EU-GDPR) \cite{regulation2016regulation}, USA's California Consumer Privacy Act (CCPA) \cite{pardau2018california}, China's Personal Information Protection Law (PIPL) \cite{feng2019future}, among others, call for high demand on safe FR to eliminate the misuse and leakage of facial data. These legal bodies make an effort to ascertain that the face appearances are both visually hidden from unauthorized viewing, and challenging for malicious attackers to retrieve at different phases of FR.

\vspace{-0.2cm}
\begin{table}[!htb]
	\scriptsize
	\caption{popular public face datasets.
	}
	\label{tab:dataset_large}
	\begin{tabular}{llcc}
		\hline
		\multicolumn{1}{c}{\textbf{Dataset}}                                                           & \multicolumn{1}{c}{\textbf{Images}}                                                                                            & \multicolumn{1}{c}{\textbf{\begin{tabular}[c]{@{}c@{}}Total\\ size\end{tabular}}} & \multicolumn{1}{c}{\textbf{\begin{tabular}[c]{@{}c@{}}Average \\ image \\ size \end{tabular}}}                 \\ \hline
		VGGFace2 \cite{cao2018vggface2}                                                                   & \begin{tabular}[c]{@{}l@{}}3.31M images of 9,131 \\ people \end{tabular}                                                                                                  & $\sim$35 GB      & 10-20 KB              \\
		\begin{tabular}[c]{@{}l@{}}IMDB-WIKI\\ \cite{pavlichenko2021imdb}  \end{tabular}                                                                & \begin{tabular}[c]{@{}l@{}}500K+ images of celebrities\\ from IMDb and Wikipedia,\\ labeled with age and gender\end{tabular} & $\sim$100 GB         & $\sim$200 KB           \\
		\begin{tabular}[c]{@{}l@{}}MS-Celeb-1M\\ \cite{guo2016ms}  \end{tabular}                                                            & \begin{tabular}[c]{@{}l@{}}$\sim$10M images of 100,000 \\celebrities \end{tabular}                                                                                       & $\sim$100 GB    & 50-150KB                \\
		\begin{tabular}[c]{@{}l@{}}Google's Open \\Images Face \\Subset \cite{amos2016openface}\end{tabular} & \begin{tabular}[c]{@{}l@{}}Part of a larger Open Images \\dataset with millions of label-\\ed objects\end{tabular}               & \multicolumn{1}{l}{$\sim$20 GB} & 50-200 KB \\
		\begin{tabular}[c]{@{}l@{}}Asian Face \\Dataset (AFD) \\\cite{xiong2018asian} \end{tabular}       & \begin{tabular}[c]{@{}l@{}}165K images of East Asian\\ faces  \end{tabular}                                                                                              & \multicolumn{1}{l}{$\sim$2.3 GB} & \multicolumn{1}{l}{$\sim$15 KB} \\
		\begin{tabular}[c]{@{}l@{}}CASIA-WebFace \\\cite{muhammad2021casia} \end{tabular}                                                            &\begin{tabular}[c]{@{}l@{}} 500K images of 10,575 \\individuals \end{tabular}                                                                                             & $\sim$2.8 GB  &    5-10 KB                \\
		FaceScrub \cite{ng2014data}                                                                  & 100K images of 530 actors                                                                                                      & $\sim$1.5 GB & \multicolumn{1}{l}{$\sim$15 KB}                    \\
		CelebA \cite{zhang2020celeba}                                                                     & \begin{tabular}[c]{@{}l@{}}200K+ celebrity images with \\40 attribute labels\end{tabular}                                     & $\sim$1.4 GB  & 50-100 KB               \\
		\begin{tabular}[c]{@{}l@{}}WIDERFACE \\ \cite{yang2016wider} \end{tabular}                                                                & \begin{tabular}[c]{@{}l@{}}32,203 images and 393,703 \\labeled faces.\end{tabular}                                            & $\sim$1.5 GB & 50-150 KB                    \\
		LFW \cite{jalal2017lfw}                                                                        & 13K+ images of 5,749 people                                                                                                    & $\sim$200 MB         & 50-100 KB           \\ \hline
	\end{tabular}
\end{table}

\vspace{-.2cm}
The increasing adoption of FRS has led to the creation of large facial datasets, and the availability of these datasets reciprocally enhances the performance and scalability of FR algorithms. Table \ref{tab:dataset_large} delineates information regarding various prominent public face datasets, including their estimated sizes. Most of the time, dataset development adhering to established regulations and standards are challenging \cite{gururaj2024comprehensive}. Even with datasets constructed in adherence to legal and ethical standards, the user's inherent right to control their personal data, remains challenging to enforce.

A notable challenge in the facial datasets is their extensive replication. Replication happens when different organizations seek to train the FR models, each necessitating an independent copy of the substantial datasets present in their respective repositories. The growth of these large datasets, along with their replication across various workstations, not only consumes considerable storage resources, but also presents extreme difficulties in maintaining user control over their own personal data.

Currently, when individuals contribute with their facial data, they relinquish control on how their information is utilized. There is no mechanism to enforce the \textit{Right-To-Be-Forgotten} (RTBF) property, which allow users to withdraw consent and prevent their face from being used in future training, even when current regulations, data protection and treatment policies are obliged to include rules for the deletion of personal data once a person withdraws their consent. Now, once the data has been incorporated into a dataset, any attempt to enforce this property is impossible. Due to this, the individual's ability to manage their personal information is undermined, and they are left with no recourse to ensure that their data is not perpetuated or utilized in ways that they may no longer approve of. As a consequence, the ethical and legal challenges associated with data ownership and user consent have not been resolved, which highlights a significant gap in the existing frameworks.

Another challenge in FR arises from Model Inversion (MI) attacks. MI attacks are first coined in \cite{MI_attacks}, and are less examined in FR domain. MI attacks present a substantial risk to the security and privacy of FRS by exploiting weaknesses in ML models to reconstruct facial images utilized in training \cite{dibbo2023sok}. These attacks utilize the model's output, including confidence scores or embeddings, to reverse-engineer and approximate the original input data. The viability of such reconstructions presents significant privacy issues, as it may reveal sensitive facial information. Attackers may exploit this vulnerability to obtain private face images used in a FR model training. MI attacks pose challenges to compliance with data protection regulations, including the GDPR \cite{regulation2016regulation}. The absence of strong protections against MI attacks reveals a significant deficiency in the technical and regulatory frameworks overseeing FR technologies, requiring immediate actions to address these vulnerabilities.

Considering all these, this paper introduces a novel framework called VOIDFace, with secure training data storage and a distributed patch-based training mechanism with the help of Visual Secret Sharing (VSS) \cite{naor1995visual}. Here, VOIDFace framework is divided into two parts, the former proposes a novel mechanism to store training data, and the latter shows a secure distributed training framework using the stored data. Thus, the VOIDFace presents a four-fold contribution:

\vspace{-0.2cm}
\begin{enumerate}
	\item VOIDFace is the first method which tries to solve the problem of training data replication issue.
	\vspace{-0.2cm}
	\item VOIDFace is also the first method which introduces the property called RTBF, which allows users to decide whether their face can be used for further training and provide ultimate control over their data.
	\vspace{-0.2cm}
	\item VOIDFace introduces a privacy preserving patch based distributed training mechanism which can efficiently use the proposed storage framework.
	\vspace{-0.2cm}
	\item VOIDFace uses training with secret shares instead of whole face image, making the FR training safer and secure against MI attacks.
	
\end{enumerate}
\vspace{-0.2cm}

Section \ref{sec:RW} briefly explains different related works. Detailed explanation of different phases in VOIDFace is explained in section \ref{sec:sysD}. The experiments results and discussion are explained in section \ref{sec:randd}, followed by conclusion in section \ref{sec:con}. 

\vspace{-0.1cm}
\section{Related works}
\label{sec:RW}
\vspace{-0.1cm}
\subsection{Privacy preserving FR}
\vspace{-0.1cm}

Privacy-preserving FR techniques aim to authenticate or identify individuals while protecting their facial data from misuse. These methods often employ encryption, such as homomorphic encryption \cite{yang2022design}, which allows computations on encrypted data without decryption, ensuring sensitive facial features remain secure. Another approach is federated learning \cite{woubie2024maintaining}, where models are trained across decentralized devices without sharing raw data, thus minimizing privacy risks. Techniques like Secure Multi-Party Computation (SMPC) also enable collaborative face matching without exposing biometric templates \cite{sayyad2020privacy}. Additionally, differential privacy \cite{chamikara2020privacy} can be applied to add noise to datasets, preventing the re-identification of individuals while maintaining recognition accuracy.

\vspace{-0.1cm}
\subsection{Federated learning}
\vspace{-0.1cm}

Federated learning is a decentralized machine learning framework that enables multiple entities or devices to collaboratively train a shared model while keeping data localized. This approach is particularly valuable in scenarios where data privacy, security, and regulatory compliance are paramount, such as in healthcare, finance, biometrics, etc. By storing data locally and sharing only model updates, federated learning minimizes the risk of data breaches, enhances privacy and security, and ensures compliance with regulations like the EU-GDPR \cite{regulation2016regulation}. Adhering to the principles defended by different regulations such as the EU-GDPR, is critical for any technology handling personal data, including federated learning systems. Several studies have explored how FL can be designed to meet these regulatory requirements \cite{privacyGDPR}. Additionally, federated learning reduces the need for large-scale data transfers, conserving bandwidth and computational resources. 

\vspace{-0.1cm}
\subsection{Visual Secret Sharing}
\vspace{-0.1cm}

Visual Secret Sharing (VSS) is a cryptographic method that partitions visual information into multiple shares, each of which appears as independent random noise images, yet collectively discloses the original image when superimposed \cite{vss}. The primary benefits of VSS include: 1. improved security, since individual shares disclose no information, 2. ease of reconstruction without intricate calculations, merely stacking the shares, 3. avoid use of key for both encryption and decryption, and 4. versatility, as it is compatible with both digital and printed formats, rendering it exceptionally adaptable for secure communication. Thus, VSS is extensively used in different state-of-the-art applications, such as medical data security \cite{mohammed2024tamper}, resource constrained applications \cite{bachiphale2024comprehensive} and many more \cite{parihar2024survey}. 

In VSS, the access structure defines the rules to determine the qualified and forbidden sets required for collaborative reconstruction of the hidden images \cite{sasaki2014formulation}. In a $(k,n)$VSS, where $n$ represents the total number of shares, and $k$ is the minimum number of shares needed for reconstruction, any subset of shares with size $\ge k$ is considered as a qualified set and otherwise forbidden set. For a multiple image VSS, the access structure is defined as follows:

\textbf{Access structure:} Let $S$ represent a finite set of shares, with $m$ $\in$ $\mathbb{N}$ (where $\mathbb{N}$ denotes the set of natural numbers) hidden images. For $i$ $\in$ $\left\lbrace1, 2, \dots, m\right\rbrace$, let $Q^i$ and $F^i$ be subsets of the power set $2^S$, such that $Q^i$ $\cap$ $F^i$ = NULL. The access structure $\Gamma^m=\left\{\left(Q^i, F^i\right)\right\}_{i=1}^m$ for $m$ hidden images (secrets) is valid if monotonicity (Eq. \ref{eq:mono1}) holds for $Q^i$ and $F^i$.

\vspace{-.4cm}
\begin{equation}
	\label{eq:mono1}
	\footnotesize
	\begin{split}
	A \in Q^i \quad \land \quad A \subseteq B \quad \Rightarrow \quad B \in Q^i \\
	B \in F^i \quad \land \quad A \subseteq B \quad \Rightarrow \quad A \in F^i
	\end{split}
\end{equation}
for all $A$, $B$ $\subseteq$ $S$ and i $\in$ $\left\lbrace 1, 2, \dots, m\right\rbrace$, and uniqueness (Eq. \ref{eq:uniq}),
\begin{equation}
	\footnotesize
	\label{eq:uniq}
	i \neq j \Rightarrow \quad \left(Q^i\right)_0  \cap \left(F^j\right)_0 = NULL 
\end{equation}
for all $i$, $j$ $\in$ $\left\lbrace 1, 2, \dots, m\right\rbrace$, and $\left(Q^i\right)_0$, $\left(F^j\right)_0$ represents the minimal element of $Q^i$ and $F^j$, respectively. Here, $Q^i$ and $F^j$ represent the qualified and forbidden set of $i^{th}$ and $j^{th}$ secret, respectively. In an access structure, if every subset of the shares are included in either $Q$ or $F$, then it is called perfect access structure. A perfect access structure can be represented using the qualified set $Q$ ($F$ = $2^S-Q$). Generally, the qualified set is represented using its minimal elements $\left(Q\right)_0$. 

\vspace{-0.1cm}

\section{System design}
\label{sec:sysD}
\vspace{-0.1cm}

\begin{figure*}[t]
	\centering
	\includegraphics[width=1\linewidth, height=9cm]{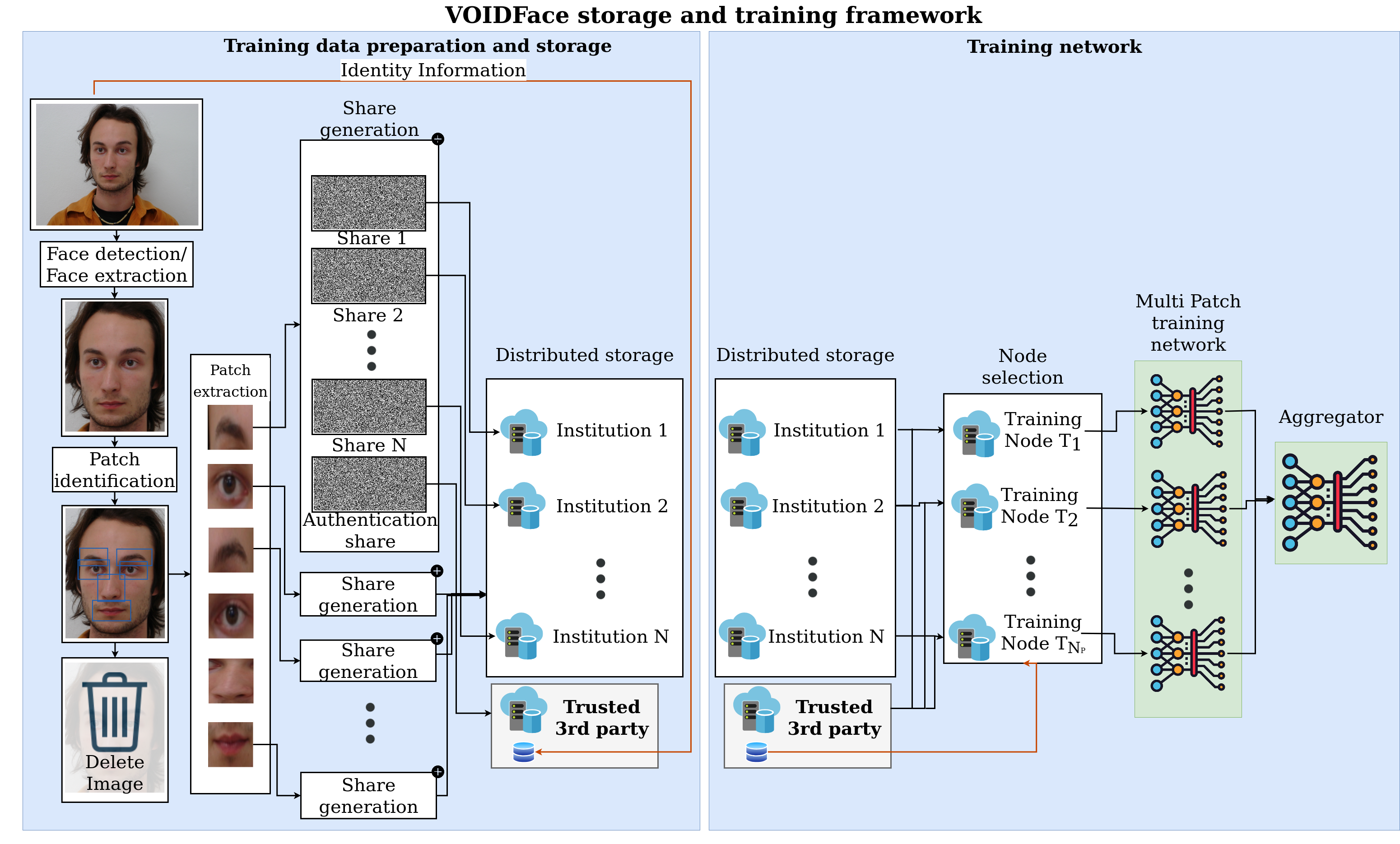}
	\caption{The block diagram of VOIDFace framework with patch identification, extraction, share generation and distribution is illustrated in privacy preserving training data preparation and storage (left), and node section, patch training and aggregator is illustrated in privacy preserving training network (right).}
	\label{fig:proposed}
	\vspace{-0.4cm}
\end{figure*}

This section provides a comprehensive overview of VOIDFace storage and training system design. VOIDFace is the first work that tries to solve the problem of training data replication, provide data control with RTBF property, and with secure privacy preserving training. Hence, the system is built on the following assumptions,

$Assumption$: Only front faces are considered.

$Justification$: Front facing images are common in the case of numerous verification and identification applications such as, immigration, ID verification, and many more. Front facing images provide a clear, unobstructed view of the face, which is essential for the FR model to learn key facial landmarks such as the eyes, nose, mouth, and overall facial structure. Moreover, these key facial landmarks have a significant contribution in VOIDFace framework.

$Assumption$: Presence of a trusted third party.

$Justification$: Conventional privacy preserving FR focuses on security of FR models. Whereas, VOIDFace focuses on both secrecy of training data and privacy preserving training. Hence, a preprocessing phase is mandatory to safeguard and store the training data after the acquisition. The presence of an entity is mandatory to initiate such preprocessing phase. Therefore, the inclusion of a trusted third party or server is essential in this design.

$Assumption$: Availability of $N$ data storage units and $N_p$ training nodes.

$Justification$: As VOIDFace uses distributed mechanism (like federated learning) for secure training data storage and patch based training, $N$ data storage units and $N_p$ training workstations are needed. The values of $N$ and $N_p$ are selected according to the number of available storage unit, and number of extracted landmarks from a face image, respectively.

Figure \ref{fig:proposed} presents the block diagram of VOIDFace, with one block depicting the training data preparation and storage, and another illustrating the multi-network training. Detailed explanation of these stages are as follows.

\vspace{-0.1cm}
\subsection{Training data preparation and storage}
\vspace{-0.1cm}

In conventional FR, the training begins after face acquisition and alignment. However, in VOIDFace, as the security of training images forms a significant contribution, a preprocessing step is performed before training. A trusted third party is responsible for the entire data preparation and storage phase. This phase is divided into three steps, patch extraction, share generation, and share distribution. A detailed description of these steps are as follows.

\textit{Patch extraction}: The first step in VOIDFace training data preparation and storage phase involves detecting and segmenting the front facial region acquired from the user. By isolating the face from the background and other irrelevant areas, the system focuses exclusively on facial features essential for recognition. This minimizes exposure to the full facial image, enhancing privacy.

In VOIDFace, once the facial region is extracted, the second step involves identifying and extracting $N_p$ specific privacy-preserving landmarks or patches. These patches are meticulously selected to contain sufficient distinguishing information for accurate recognition, while avoiding reliance on the entire facial image. After patch extraction, the full face image is permanently deleted and never reconstructed at any stage within the VOIDFace framework. By relying on the extracted $N_p$ patches, the system achieves reliable FR without storing or processing the complete face.

\textit{Share generation}: After extracting all $N_p$ privacy-preserving facial patches, these patches are resized uniformly and processed through the VSS module for share generation. Since each user has $N_p$ patches, a multiple patch secret sharing scheme is adapted to include all $N_p$ patches for share generation.

In VOIDFace, each user has $N_p$ equally sized patches, which are treated as $N_p$ multiple secrets for share generation. Initially, a $1$-$out$-$of$-$N_p$ patch (denoted as $P_1$) is encrypted into two random grids using a random number generator function. One of these random grids serves as the private share ($PS_1$) for $P_1$, while the other serves as the authentication share ($AS$). Thus, for $i$ $\in$ $\left\lbrace 2,3,\dots,N_p\right\rbrace$, the private share $PS_i$ is generated as follows:
\vspace{-.1cm}
\begin{equation}
	\footnotesize
	PS_i = P_i \oplus AS
\end{equation}

\vspace{-.1cm}
where $AS$ represents the authentication (common) share, each $P_i$ corresponds to the $i^{th}$ patch, with $i$ $\in$ $\left\lbrace 2,3,\dots,N_p\right\rbrace$, and $\oplus$ denotes the logical XOR operation.

VOIDFace framework uses a perfect access structure with single set in qualified set $Q$, generally known as minimally refined perfect access structure. The VOIDFace access structure $\Gamma^{N_p}=\left\{\left(Q^i, F^i\right)\right\}_{i=1}^{N_p}$ for $N_p$ patches is formulated as,

\vspace{-.3cm}
\begin{equation}
	\footnotesize
	\begin{split}
	\left(Q^i\right)_0 = \left\lbrace \left\lbrace AS, \ PS_i \right\rbrace \right\rbrace \\
	\left(F^i\right)_0 = 2^S - \left(Q^i\right)_0
\end{split}
\end{equation}

\textit{Share distribution}: Once the $AS$ and private shares are generated for each individual, the $AS$ is stored with a trusted third party to support the user’s RTBF property. Private shares are distributed across various institutions to enable security and distributed storage mechanism. As private shares are stored in institutions instead of complete faces, VOIDFace eliminates the data replication problem and introduces additional security and privacy \cite{chou2024secure}, and also mitigates the risk of single-point attacks. Consequently, each storage location holds only one or a limited number of shares, and thereby increasing the security.

Here, a situation can arise in which the number of private shares ($N_{ps}$) and number of available storage institutions ($N$) are different. In VOIDFace, this situation is handled as follows,

\textit{Case 1:} $N_{ps} > N$, In this case, VOIDFace randomly chooses $N$ private shares for further processing.

\textit{Case 2:} $N_{ps} < N$, In this case, first calculate \( j = N - N_{ps} \), if $j \le N_{ps}$, then randomly selected $j$ private shares are encrypted into two random grids. Similarly if $j > N_{ps}$, then randomly selected $j$ private shares are encrypted into multiple random grids to make total $N$ random grids. In such a scenario, the private shares of $P_i$ consist of all the random grids generated from the associated private share $PS_i$. This expansion supports on-demand scalability of the VOIDFace framework to accommodate the participation of more institutions.

\vspace{-0.1cm}
\subsection{Privacy preserving training}
\label{sec:fe}
\vspace{-0.1cm}

After securing the training data, the privacy-preserved distributed data is now ready for model training. This privacy preserving training is inspired on federated learning \cite{wei2020federated}. In VOIDFace, different stages in privacy preserving training include, node selection and dropping, patch reconstruction, patch training, and aggregation.


\textit{Node selection and dropping}: During node selection, a set of institutions/workstations are selected for patch reconstruction and training. Even though there are different strategies of node selection such as, random, resource based, correlation based, energy-aware, noise-aware, and many more, we choose a resource based selection called FedCS \cite{nishio2019client} for VOIDFace. The main reason for this selection is due to its resource aware selection, where the client selection is based on their resource conditions, such as computational power, bandwidth, and energy availability. FedCS is designed to handle dynamic scenarios where client resources may fluctuate. This is particularly useful in a distributed environment, where participating institutions may have varying levels of availability and performance as in VOIDFace. FedCS aims to minimize communication overhead by selecting clients that can complete the training tasks within a reasonable time frame. This is crucial for VOIDFace, which require frequent communication between storage facilities. FedCS is scalable and can handle a large number of clients, making it suitable for environments where facial data is distributed across multiple storage locations. VOIDFace also uses node removal strategy introduced  by Huang et al. \cite{huang2022stochastic} called E3CS. In VOIDFace, when some nodes may become unavailable or slow, E3CS can dynamically adjust by removing these nodes from the training process, with faster convergence, ensuring fairness and robustness.

\textit{Patch reconstruction}: In VOIDFace, when FR training is requested, the requester must first communicate with a trusted third party. Before validating the request, the trusted third party verifies user authorization to access the user's facial data by checking the authentication shares. Once the authentication shares are validated, the training request and authentication shares are sent to $N_p$ independent, non-communicating selected training workstations ($T_1$ to $T_{N_p}$). Each workstation $T_i$ collects the $i^{th}$ private share from the relevant counterpart. After receiving these private shares, the corresponding patches are reconstructed on each of the $N_p$ independent, non-communicating, workstations using,

\vspace{-0.1cm}
\begin{equation}
	\footnotesize
	\overline{P_i} = \overline{AS_i} \oplus \overline{PS_i}
\end{equation}  
where $\overline{P_i}$, $\overline{AS_i}$, and $\overline{PS_i}$ are the $i^{th}$ reconstructed patch, $i^{th}$ authentication share  which is active, and $i^{th}$ private share, (or shares) respectively.  Here,  $\overline{X}$ ($\overline{P_i}$, $\overline{AS_i}$, and $\overline{PS_i}$) represents the value of $X$ during training. Ideally, $\overline{X}$ and $X$ would be same. Here, we represented $\overline{X}$ instead of $X$, as all the original images are removed after the patch extraction during the privacy preserving data preparation and storage phase, which is independent from the training phase.

\textit{Patch training and aggregation:} After reconstruction, the patches are processed by a deep learning-based FR patch based multi-network for training. VOIDFace privacy-preserving training framework includes two types of networks: \textit{Patch Training Network} (PTN) and \textit{Aggregator}. Since each user has $N_p$ distinct patches, VOIDFace utilizes $N_p$ individual PTN. Each PTN comprises parallel backbone CNNs designed to extract feature embeddings from the corresponding reconstructed patches. The \textit{Aggregator} integrates the outputs from the PTN and can be structured in various configurations. In VOIDFace, the output of the \textit{Aggregator} is an N-dimensional feature vector derived from the PTN outputs, predicting the face embedding based on the input set of face image patches. In VOIDFace, each PTN component employs MobileNet, while the \textit{Aggregator} is implemented as a fully connected layer that consolidates these outputs into a final global feature vector.

\vspace{-0.1cm}
\section{Experiment Results and Discussion}
\label{sec:randd}
\vspace{-0.1cm}
\subsection{Experimental setup}
\vspace{-0.3cm}
\begin{table}[!htb]
	\scriptsize
	\centering
	\begin{threeparttable}
		\caption{VGGFace2 dataset details.}
		\label{tab:dataset}
		\begin{tabular}{cccc}
			\hline
			\textbf{Filtering}                                            & \textbf{\begin{tabular}[c]{@{}c@{}}\# images\end{tabular}} & \textbf{\begin{tabular}[c]{@{}c@{}}\# Classes\end{tabular}} & \textbf{\begin{tabular}[c]{@{}c@{}}\# Average Images \\per Class\end{tabular}} \\ \hline
			
			\begin{tabular}[c]{@{}c@{}}None\end{tabular} & 3074k                                                               & 8631                                                                  & 356                                                                 \\
			\begin{tabular}[c]{@{}c@{}}FRR=0.05\end{tabular} & 1158K                                                               & 8628                                                                  & 134                                                                 \\
			\hline
		\end{tabular}
		\begin{tablenotes}
			\raggedleft
			\scriptsize
			\item \# - Number of
		\end{tablenotes}
	\end{threeparttable}
	\vspace{-0.2cm}
\end{table}
\vspace{-0.2cm}
\textit{Datasets:} Data preparation and storage is an integral part of the VOIDFace framework. Currently, preparing new data from scratch is challenging due to different issues like, privacy and consent, bias and diversity limitations, high volume requirements, and regulatory compliance. Therefore, we currently rely on an existing dataset, VGGFace2 \cite{cao2018vggface2}. Since VOIDFace uses only front facing images (see assumptions in section \ref{sec:sysD}), and experiments are conducted on an existing dataset instead of a proprietary dataset, a filtering mechanism is adopted to select only front facing images. As VGGFace2 is an uncontrolled and extensive dataset, front facing images are selected through a quality-driven filtering approach presented in \cite{medvedev2023improving}. This filtering uses low False Rejection Rate (FRR) of 0.05 to select front facing high-quality images. A low FRR threshold reduces the exclusion of genuine images by emphasizing clear, straightforward, front facing poses. Further details regarding the dataset is given in Table \ref{tab:dataset}.

\textit{Preprocessing:} Preprocessing is an essential part in VOIDFace training data preparation and storage. During VOIDFace preprocessing, the user's face is detected and segmented to isolate relevant regions. We have used Histogram of Oriented Gradients (HOG)-based method combined with linear classifier to detect and segment frontal face images. This technique offers high accuracy and efficiency for real-time applications without requiring an additional model. For the experimentation, six patches ($N_p$= 6) are used including both eyes, both eyebrows, nose, and mouth. Thus, during preprocessing, the frontal images and six patches are extracted using a pre-trained model\footnote[1]{\url{https://www.kaggle.com/code/zeyadkhalid/face-landmarks-detection-and-alignment-dlib}}.

\textit{Models:} VOIDFace training phase combines two types of networks for privacy-preserving training: \textit{Patch Training Network} (PTN), and \textit{Aggregator}. PTN is built on a MobileNet backbone and concludes with an output dense patch feature layer. Features extracted from different patches are concatenated and fed into the \textit{Aggregator}. Here, PTN is optimized using the SGD optimizer with a momentum of 0.5, and an initial learning rate of 0.01, which follows a Cosine Annealing schedule ($\eta_\text{max}$ = 0.01, $\eta_\text{min}$ = 1e-7) over 20 epochs. Training uses a batch size of 10 and processes 96×96 RGB images of all the six facial patches (two eyebrows, two eyes, nose, and mouth). Images are normalized by subtracting [0.5, 0.5, 0.5] and scaling by 1/255. We employ categorical cross-entropy loss with equal weights of 1.0 for all patches, trained on a dataset of 8628 classes. Key architecture parameters include an alpha of 1.4 (width multiplier) and an ArcFace margin of 0.5. Similarly, the \textit{Aggregator} consists of a simple fully connected layer that consolidates the concatenated patch feature outputs into a global feature vector as final output.

Additionally, we also explored a modified version of PTN, incorporating extra classification supervision at the patch level, transforming the problem into a multitask learning approach where both PTN and \textit{Aggregator} perform the same classification task. This approach enhances patch-level supervision and offers greater flexibility in patch selection and augmentation. In further experiments, we call the model without supervision - V1 and the model with supervision - V2.

\vspace{-0.1cm}
\subsection{Training dataset analysis}
\vspace{-0.4cm}

\begin{figure}[!htb]
	\centering
	\includegraphics[width=1\linewidth]{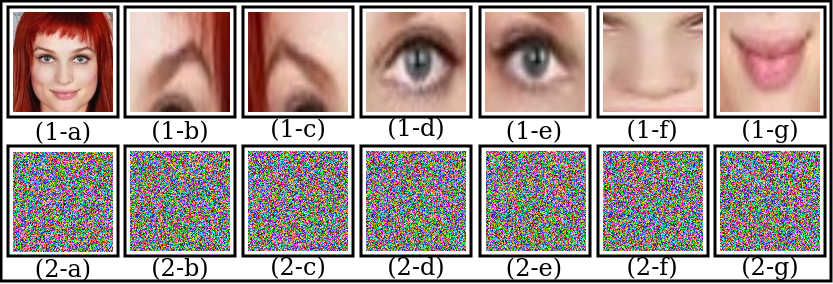}
	\caption{A sample face image, six face patches and corresponding shares.}
	\label{fig:Share}
	\vspace{-0.3cm}
\end{figure}

In VOIDFace, as each institution gets only a secret share generated from an extracted patch, getting any information about the original face or even the corresponding patch is impossible for the participating institutions. Figure \ref{fig:Share} presents a sample face image (Figure \ref{fig:Share}(1-a)), the extracted patches (Figures \ref{fig:Share}(1-b) to \ref{fig:Share}(1-g)) and the generated secret shares (row 2). Here, Figures \ref{fig:Share}(2-b) to \ref{fig:Share}(2-g) show the private shares corresponding to Figures \ref{fig:Share}(1-b) to \ref{fig:Share}(1-g), respectively. Figure \ref{fig:Share}(2-a) shows the authentication share, which is common to all patches. All of these shares are meaningless and provide no visual information about the original face image or the corresponding patch. More details on information content within the share is presented in supplementary material (Section 1.1). Hence, VOIDFace protects the  privacy of the individuals while preparing the training data. 

Training data replication is another challenge in existing FR systems. Currently, if two institutions want to train a FR system using the same dataset, a replica of the whole dataset needs to be stored in each institution. This situation increases the storage requirement, increases the complexity in managing data, and violates privacy and security principles. In VOIDFace, since each institution gets only shares generated from a patch instead of entire image, there is considerable reduction in the storage requirement. The average image size of popular public face datasets (compressed version) varies from 50-200kB (see Table \ref{tab:dataset_large}). Uncompressed or high quality face datasets may contain bigger size images, even with average image size in Mega Bytes (MBs). In these cases, replicating the training data at different institution needs large storage requirement. Whereas in VOIDFace, each share are sized $\le$10kB, which shows a huge reduction in storage requirement. Hence, this experiment shows that the VOIDFace framework stores face image visually concealed (as secret shares) with less storage requirements and solves data replication problem.

The RTBF, also known as the \textit{right to erasure}, is a critical data protection principle enshrined in several major privacy regulations. Article 17 of th EU-GDPR states that individuals have the right to request the deletion of their personal data when the data is no longer necessary for its original purpose, or the individual withdraws consent, or the data was unlawfully processed, or to comply with a legal obligation. Existing FR systems can technically comply with the RTBF, but they require manual intervention and human decision-making, making the process inefficient and inconsistent. Additionally, when a dataset is replicated in several institutions, which is very common, the need to request these institutions to delete the withdraw data implies a very high risk of these institutions disregard the deletion request. VOIDFace automates this procedure, offering a first practical and scalable solution to effectively enforce the RTBF property by allowing individuals to remove their face information from further training. In VOIDFace, this property is fulfilled with the help of a trusted third party and authentication share ($AS$). The trusted third party validates the $AS$ to confirm the activation of user's participation in VOIDFace training. Any user who wish to not participate in further training can request trusted third party to remove the corresponding $AS$. Without $AS$, none of the face patches corresponding to the particular subject can be used for further FR training. This provides the user RTBF property to control their own face data. Since the deletion of $AS$ can be done immediately upon getting the request, the RTBF operation is computationally efficient. 

Although this strategy is computationally efficient and improves the control, it also creates a new issue, which is the existence of abandoned shares. Abandoned shares are the private shares with no authentication shares, and are useless. In VOIDFace, this problem is solved using a simple and effective solution by periodically checking $AS$, and discarding the abandoned private shares.

\vspace{-0.1cm}
\subsection{Security analysis}
\vspace{-0.1cm}

This section explains the possibility of different attacks on VOIDFace framework.

\textit{Brute-force attack:} The security of cryptographic techniques heavily depends on the secrecy of keys. The absence of keys in VOIDFace is advantageous and enhances security. Therefore, brute-force attacks in VOIDFace are estimated based on the ability to guess the VSS share and the share correlation coefficient in all directions. 

In VOIDFace training data storage phase, as shares are generated using a randomized strategy, each pixel takes a value 0-255 across three color channels. This results in a probability of $\frac{1}{256}$ to guess a pixel. For a share of size $w \times h \times 3$, the probability becomes $(\frac{1}{256})^{w \times h \times 3}$. As in this experimentation, for a share of size 96$\times$96$\times$3, this probability would be $9.581622535 \times 10 ^ {-66584}$, a negligible value. Even with this, an attacker would only be able to guess a single share. To reconstruct the entire patch, multiple shares are required, which further increases the complexity exponentially. We also performed an experiment to evaluate encryption strength using correlation coefficient. Details are available on supplementary material (Sec 2.2). Given this large sample space and low correlation coefficients, a brute-force attack on VOIDFace is virtually impossible.

\textit{Statistical attack:} In any secret sharing based application, generating different share each time is essential as it prevents predictability and brute-force attack, stops cheating and share usage, and protects from statistical analysis. Due to randomness in VOIDFace, each time the shares generated from the same patches are different to each other. To validate this, we have used a matrix called Number of Pixel Change Rate (NPCR) to evaluate the pixels change. NPCR also shows the resiliency towards differential attacks. Higher values of NPCR shows more resilient whereas lower shows opposite. More details on NPCR is given in \cite{wu2011npcr}. 

In this experiment, each patch is encrypted 1,000 times, and the NPCR is calculated by comparing a randomly selected share with all the remaining 999 shares. Table \ref{tab:npcr} presents the average NPCR values calculated per patch. These high values ($>$98\%) show a strong indicator of robust encryption, demonstrating high sensitivity, uniform diffusion and protection from statistical attacks.

\begin{table}[!htb]
	\centering
	\scriptsize
	\caption{NPCR values for 1,000 distinct shares per patch.}
	\label{tab:npcr}
	\begin{tabular}{cccccc}
		\hline
		\begin{tabular}[c]{@{}c@{}}Left \\ eyebrow\end{tabular} & \begin{tabular}[c]{@{}c@{}}Right \\ eyebrow\end{tabular} & \begin{tabular}[c]{@{}c@{}}Left \\ eye\end{tabular} & \begin{tabular}[c]{@{}c@{}}Right \\ eye\end{tabular} & Nose & Mouth \\ \hline 98.87\%  & 98.88\% & 98.69\%  &  98.75\% & 98.54\%    &  98.67\%     \\ \hline
	\end{tabular}
	\vspace{-0.4cm}
\end{table}

\textit{Model Inversion (MI) attacks:} In this section, we analyzed the resiliency of VOIDFace on MI attacks. We have used the technique presented in\cite{nguyen2023re} to simulate MI attacks. We have used VOIDFace aggregator model as the target model and face.evoLve as the evaluation model trained using CelebA dataset \cite{zhang2020celeba}. Here, the impact of MI attack is evaluated using attack accuracy (Attack Acc) and K-Nearest Neighbors Distance (KNN Dist).

\vspace{-.2cm}
\begin{table}[!htb]
	\centering
	\scriptsize
	\caption{MI attack results.}
	\label{tab:MI}
	\begin{tabular}{ccc}
		\hline
		\textbf{Method} & \textbf{Attack Acc} & \textbf{KNN Dist} \\ \hline
		ArcFace         & 82.4\%              & 1247.28           \\
		VOIDFace        & 12.1\%              & 2240.30            \\ \hline
	\end{tabular}
\end{table}
\vspace{-0.3cm}
Table \ref{tab:MI} shows the comparison of VOIDFace with Arcface \cite{deng2019arcface} against a black-box MI attack. Under a query-based attack scenario, the ArcFace model leaks private training data with 82.4\% attack accuracy, and lower KNN Dist. In contrast, VOIDFace reduces attack accuracy to 12.1\%, and high KNN Dist. These results confirm that in VOIDFace, the recovered samples deviate significantly from private data. These results demonstrate VOIDFace’s effectiveness in preserving privacy while maintaining model utility.

\textit{Protection due to distributed storage:} Another important property of VOIDFace is the distributed storage. Distributed storage enhances security by eliminating single points of failure, and reduces the attacker’s ability to compromise data. A key principle is that \textit{``an attacker cannot be omnipresent"}—they cannot simultaneously compromise all nodes in a distributed network. With distributed storage in VOIDFace, it gets computationally very hard for an attacker to get both the authentication share and private shares. Dividing and distributing private shares to many workstations again increases the complexity. Let us consider a white-box attacker who can query the VOIDFace framework and knows its detailed protection mechanism. This attacker is generally envisioned as a malicious entity wiretapping the transmission. Even with this wiretapping, the communication between the workstations transmit either an authentication share or some private share. As a subset of private shares present in the training workstations are never transmitted, this makes the reconstruction impossible. 

\begin{figure*}[!htb]
	\centering
	\includegraphics[width=0.9\linewidth]{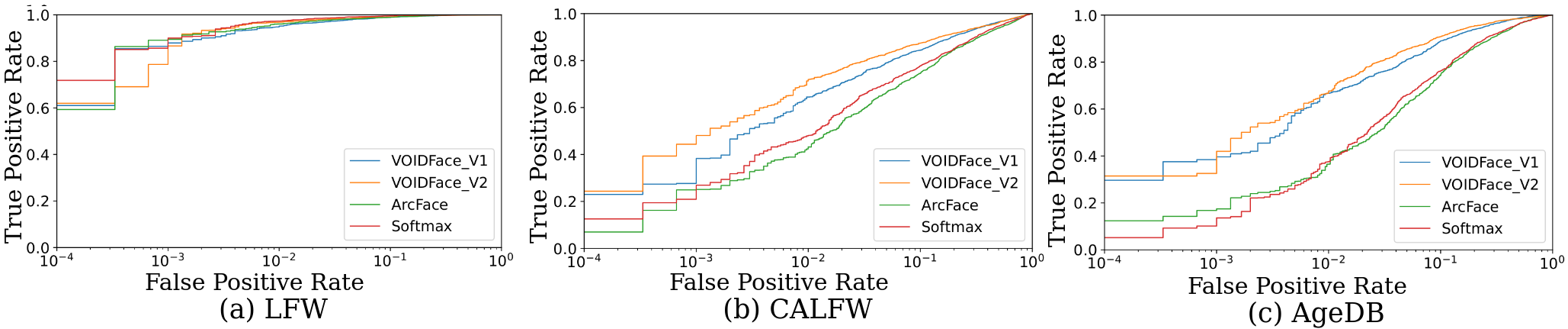}
	\caption{ROC curves of VOIDFace, Arcface, and Softmax models on various benchmarks.}
	\label{fig:performance_11}
\end{figure*}

\begin{table*}[!htb]
	\scriptsize
	\centering
	\caption{Comparison of VOIDFace with other state-of-the-art techniques.}
	\label{tab:co}
	\begin{tabular}{ccccccccc}
		\hline
		\multicolumn{1}{c}{\textbf{Technique}} & \multicolumn{1}{c}{\textbf{Used}} & \multicolumn{1}{c}{\textbf{\begin{tabular}[c]{@{}c@{}}Training data\\ Protection\end{tabular}}} &
		\multicolumn{1}{c}{\textbf{Encryption}} & \multicolumn{1}{c}{\textbf{\begin{tabular}[c]{@{}c@{}}Distributed \\ Storage\end{tabular}}} & \multicolumn{1}{c}{\textbf{RTBF}} & \multicolumn{1}{c}{\textbf{\begin{tabular}[c]{@{}c@{}}Patch-based\\ network\end{tabular}}} & \multicolumn{1}{c}{\textbf{\begin{tabular}[c]{@{}c@{}}Privacy\\ Preserving\end{tabular}}}& \multicolumn{1}{c}{\textbf{\begin{tabular}[c]{@{}c@{}}MI\\ Attacks\end{tabular}}} \\ \hline
		\multicolumn{1}{c}{\cite{yang2022design}}            & \multicolumn{1}{c}{Homomorphic Encryption}            & \multicolumn{1}{c}{No}                                                                     & \textbf{Yes} & No                                                                                            & No &  No & \textbf{Yes} & Possible                                                                               \\
		\cite{woubie2024maintaining} &   Federated learning & No  &  No & \textbf{Yes}  & No &  No &  \textbf{Yes} & Possible                                                                               \\
		\cite{chamikara2020privacy}& Differential privacy & No & No &  No & No &   No & \textbf{Yes} & Possible                                                                                 \\
		\cite{liu2016large}
		& Softmax loss                                      & No                                                                                              &  No                                                                                           &  No                                 &  No                                                                                          &  No                                                                                &  No & Possible                                                                                \\
		\cite{deng2019arcface}
		& ArcFace loss                                      & No                                                                                                &  No                                                                                           &   No                                &  No                                                                                          & No                                                                                 & No & Possible                                                                                 \\
		\multicolumn{1}{c}{VOIDFace}           & Arcface with VSS            & \textbf{Yes}                                                                      &  \textbf{Yes}                                                                                           & \textbf{Yes}                                  & \textbf{Yes}                                                                                           & \textbf{Yes}                                                                                 & \textbf{Yes} & \textbf{Challenging}                                                                                 \\ \hline
	\end{tabular}
	\vspace{-0.3cm}
\end{table*}

\vspace{-0.1cm}
\subsection{Performance analysis}
\vspace{-0.1cm}

To demonstrate the practicality of VOIDFace, we conducted a comparison with several conventional methods of training deep networks for FR. Specifically, we evaluated our model (V1 and V2) against the standard training of a single, unified deep network using Softmax loss \cite{liu2016large}, as well as its marginal modification - ArcFace \cite{deng2019arcface}. We trained both the regular Softmax and ArcFace models on the selected dataset used in the VOIDFace framework, and then tested on well-known benchmarks, including LFW \cite{LFWTechUpdate}, CALFW \cite{CALFW}, and AgeDB-30 \cite{agedb30}. These benchmarks were chosen as they do not include extreme pose variations, making a better fit for VOIDFace, which is sensitive to such factors. This comparison highlights the strengths of our method in handling typical variations, while maintaining competitive performance against traditional approaches.

The backbone for both Softmax and ArcFace (Resnet50) was selected to ensure that the number of parameters approximately matches VOIDFace to provide a fair comparison. All models were initiated with ImageNet weights and trained for 20 epochs. We utilized SGD optimizer with the learning rate linearly decaying from 0.01 to 0.00001.

Our results obtained on the chosen data subset demonstrate (Figure \ref{fig:performance_11}) that VOIDFace is effective across diverse benchmarks, proving its suitability for FR tasks. In majority of the benchmarks, V2 model (with multitask patch supervision) demonstrate better performance than V1. ArcFace provides a solid improvement in more challenging benchmarks (like CALFW and AgeDB-30) over the standard Softmax, but does not reach the robustness of VOIDFace. This comparison highlights that the patch-based approach utilized by VOIDFace contributes positively to its performance.

\vspace{-0.1cm}
\subsection{Comparison with existing techniques}
\vspace{-0.1cm}

In this section, we have compared VOIDFace qualitatively with other SOTA techniques. Here, the comparison evaluates different features such as training data protection, use of encryption, distributed storage, RTBF, use of patches, privacy preservation, and possibility of MI attacks. Table \ref{tab:co} shows the qualitative comparison of VOIDFace with SOTA techniques. Presence of all the mentioned features and challenge in MI attacks show the VOIDFace transformative potential, bridging critical gaps in data storage and training of FRS by addressing many longstanding challenges.

\vspace{-0.1cm}
\section{Conclusion}
\label{sec:con}
\vspace{-0.1cm}

Along with the popularity, FR systems face significant privacy and security challenges due to the sensitive nature of facial data. Some of the unexplored areas in FR includes secrecy of training data and protection from MI attacks, training data replication, and lack of user control. Hence, we introduce VOIDFace, a novel solution leveraging VSS to securely store and process training facial data in distributed patches, eliminating the need for full image reconstruction. This approach addresses data replication by decentralizing storage across institutions while maintaining privacy through cryptographic shares. A trusted third party manages user authorization, enabling \textit{Right-To-Be-Forgotten} by controlling access to the shares. VOIDFace also mitigates MI attacks by ensuring facial data is never stored or processed in complete form. Distributed patch-based training enhances scalability and reduces computational burdens across systems. VOIDFace’s contributions include resolving replication issues, enforcing RTBF, enabling privacy-preserving distributed training, and preventing reconstruction attacks. By combining VSS with distributed learning, it offers improved security, regulatory compliance, and user autonomy. Experimental results demonstrate its effectiveness in maintaining FR accuracy while enhancing privacy protections. Future work could explore optimization for real-time systems with better storage requirements, and broader adversarial robustness. We plan to collect and store an entire dataset using VOIDFace. VOIDFace represents a significant advancement in ethical FR development, balancing performance with stringent privacy safeguards. 

\vspace{-0.1cm}
\section*{Acknowledgment}
\vspace{-0.1cm}

This project was funded by the EU’s Horizon Europe project ACHILLES under Grant Agreement No 101189689, and from national funds through FCT - Fundação para a Ciência e a Tecnologia, under the project UID/00048.

{\small
\bibliographystyle{ieee}
\bibliography{ref}
}

\twocolumn[
\begin{@twocolumnfalse}
	\section*{\centering \Large{{Supplementary Material} -- ``VOIDFace: Towards an effective face training data storage and protection with \textit{Right to be Forgotten} Property''}}
\end{@twocolumnfalse}
]

\vspace{5em}
This supplementary material contains additional details on multi-patch training network, followed by additional experiments that provides information contents with the shares, correlation analysis and patch exclusion analysis. These experiments are conducted to reinforce the effectiveness of \textbf{VOIDFace} framework.

\section*{Additional description on multi-patch training network}

The VOIDFace multi-patch-network framework is designed around two primary components: the \textit{Patch Training Network} and the \textit{Aggregator}. This framework distributes the face recognition training task on localized facial regions (patches) further integrating their learned representations into a holistic embedding. The framework utilizes $N_s$ distinct facial patches (in our experiments, we employ $N_s = 6$). Accordingly, VOIDFace includes six independent \textit{Patch Training Network}, each specialized to process a specific facial region. Each Patch Training Network comprises a dedicated MobileNet backbone CNN that extracts discriminative feature embeddings from the input reconstructed patch. These CNNs operate in parallel, enabling the model to independently learn region-specific facial features.

The outputs of the six \textit{Patch Training Networks} (512-dimensional feature vectors) are then forwarded to the \textit{Aggregator} network. The \textit{Aggregator} is implemented as a simple fully connected layer, whose role is to integrate the patch-level embeddings into a single global feature representation of the entire face. This final embedding, also 512-dimensional, is intended to capture comprehensive identity-related information by synthesizing both local and global facial cues.

\begin{figure}[htbp]
	\centering
	\includegraphics[width=0.8\linewidth]{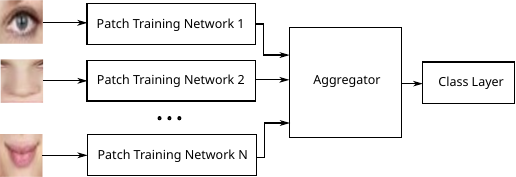}
	
	\textit{(a)} 
	
	\vspace{1mm}
	
	\includegraphics[width=0.97\linewidth]{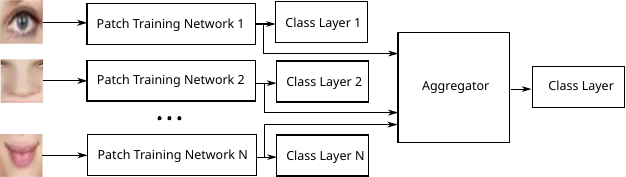}
	
	\textit{(b) } 
	\caption{Multi-Patch Training Network architecture schematics. a) - V1; b) - V2.}
	\label{fig:schematics}
\end{figure}

We investigate two architectural variants of the multi-patch-network framework (illustrated in Figure  \ref{fig:schematics}):
\begin{itemize}
	\item {\textit{Version 1 (V1)} adopts a traditional single-task learning approach. In this setup, the model is trained as a standard classification problem: the \textit{Aggregator} outputs a class prediction corresponding to the identity label. The Patch Training Networks act purely as feature extractors, and only the final integrated embedding is supervised through classification loss.}
	
	\item {\textit{Version 2 (V2)} redefines the task as a multi-task learning problem. In this enhanced configuration, additional classification heads are introduced at the output of each Patch Training Network. Consequently, both the Patch Training Networks and the \textit{Aggregator} are supervised to perform the same identity classification task. This patch-level supervision reinforces the learning of discriminative features at the local level and provides more robust training signals. }
\end{itemize}




\begin{figure*}[htbp]
	\centering
	
	\includegraphics[width=0.325\linewidth]{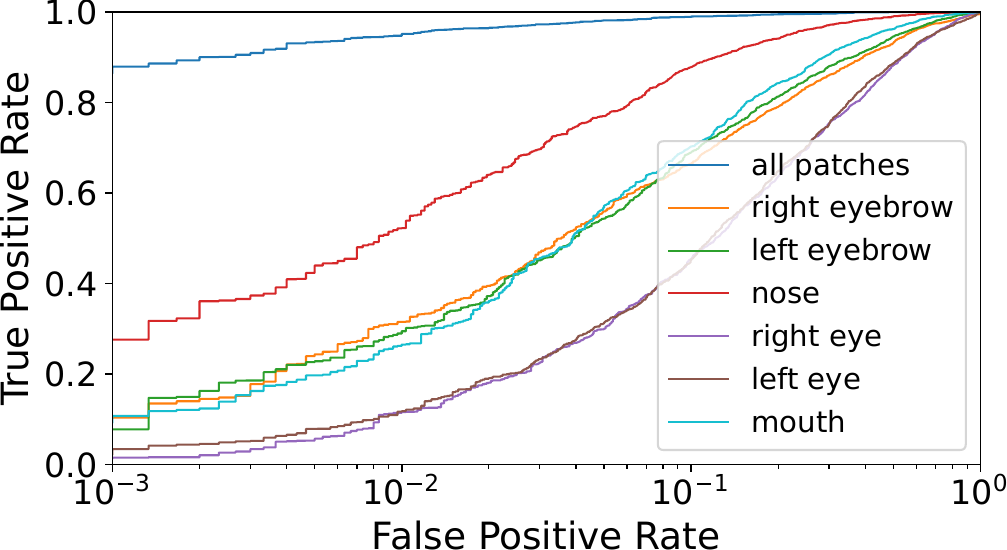}
	\includegraphics[width=0.325\linewidth]{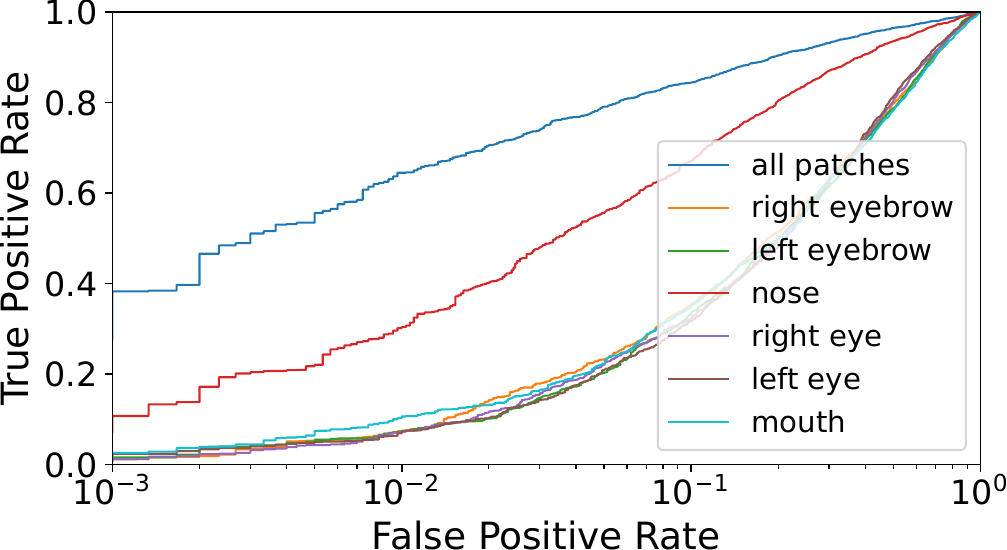}
	\includegraphics[width=0.325\linewidth]{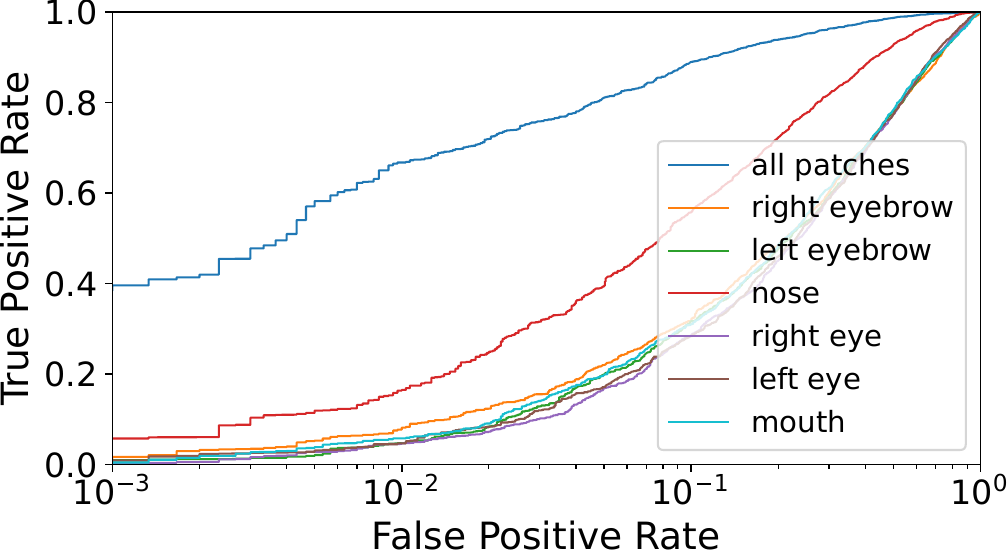}
	
	\textit{(a) LFW} \hspace{120pt}   \textit{(b) CALFW} \hspace{120pt}   \textit{(c) AgeDB}
	
	\includegraphics[width=0.325\linewidth]{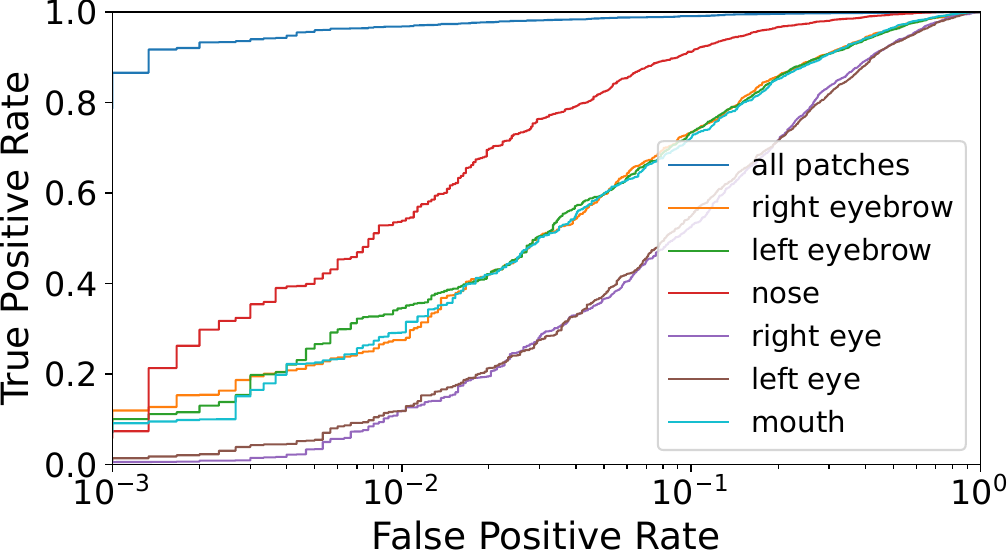}
	\includegraphics[width=0.325\linewidth]{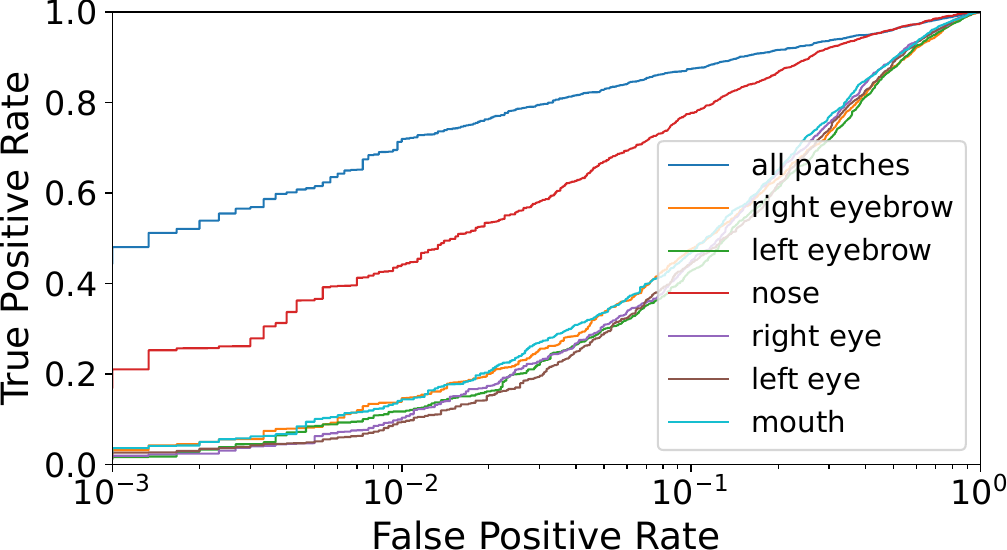}
	\includegraphics[width=0.325\linewidth]{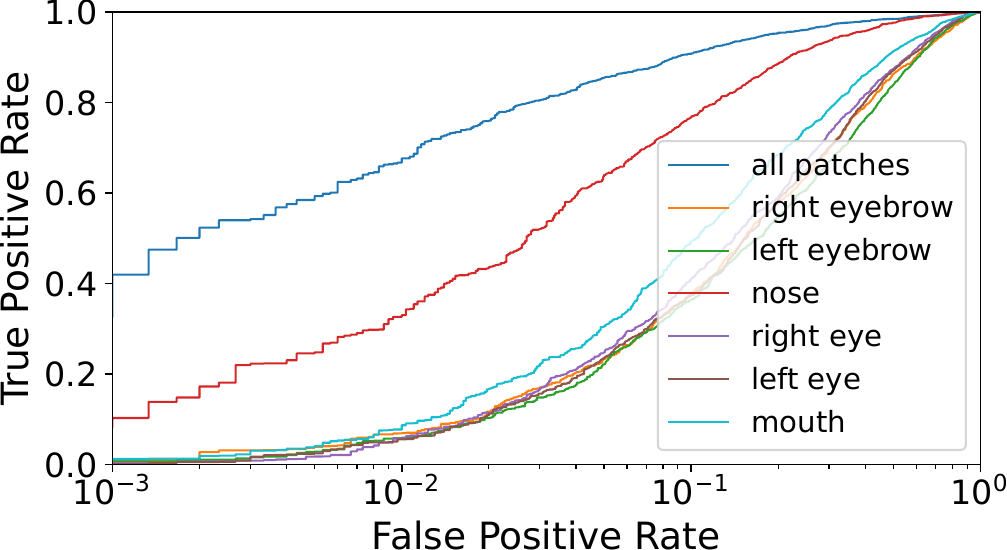}
	
	\textit{(d) LFW} \hspace{120pt}   \textit{(f) CALFW} \hspace{120pt}   \textit{(e) AgeDB}
	\caption{ROC curves of VOIDFace Single-Patch models (V1-a, b, c; V2-d, f, e) on various benchmarks.}
	\label{fig:performance_11}
\end{figure*}

\begin{figure*}[htbp]
	\centering
	
	\includegraphics[width=0.325\linewidth]{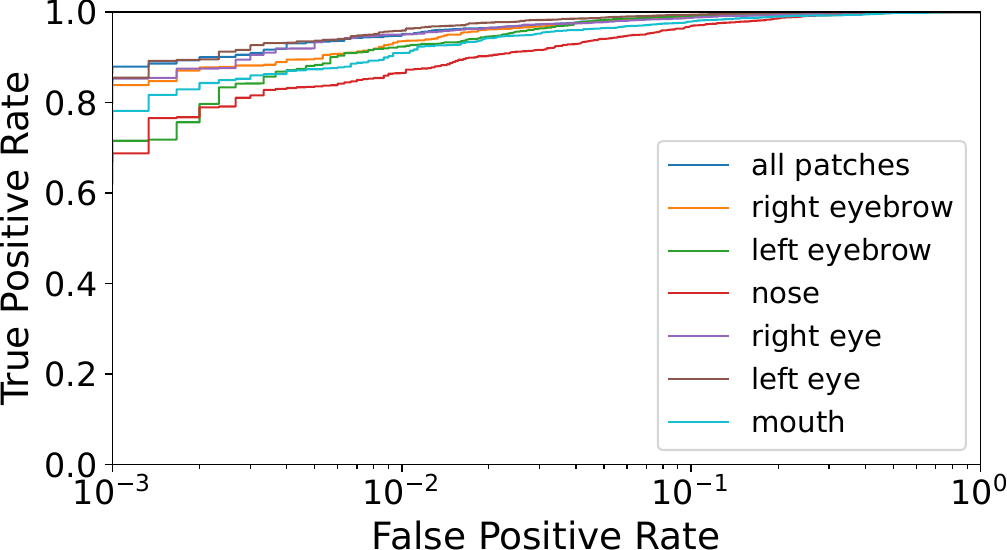}
	\includegraphics[width=0.325\linewidth]{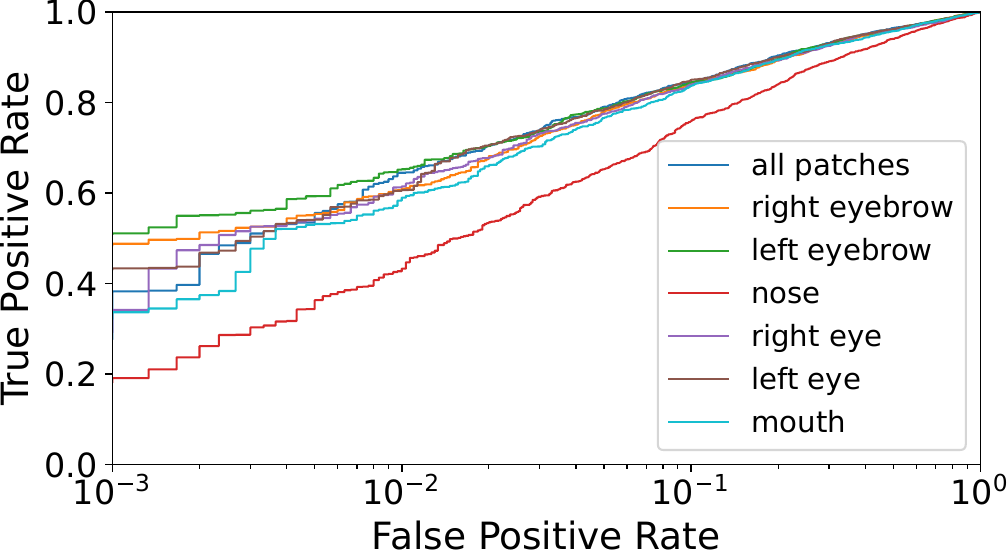}
	\includegraphics[width=0.325\linewidth]{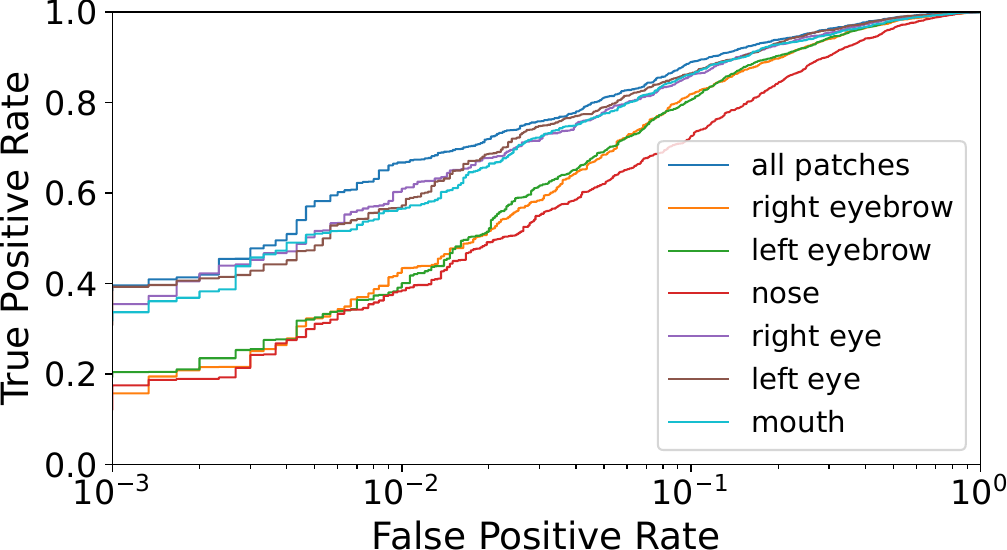}
	
	\textit{(a) LFW} \hspace{120pt}   \textit{(b) CALFW} \hspace{120pt}   \textit{(c) AgeDB}
	
	\includegraphics[width=0.325\linewidth]{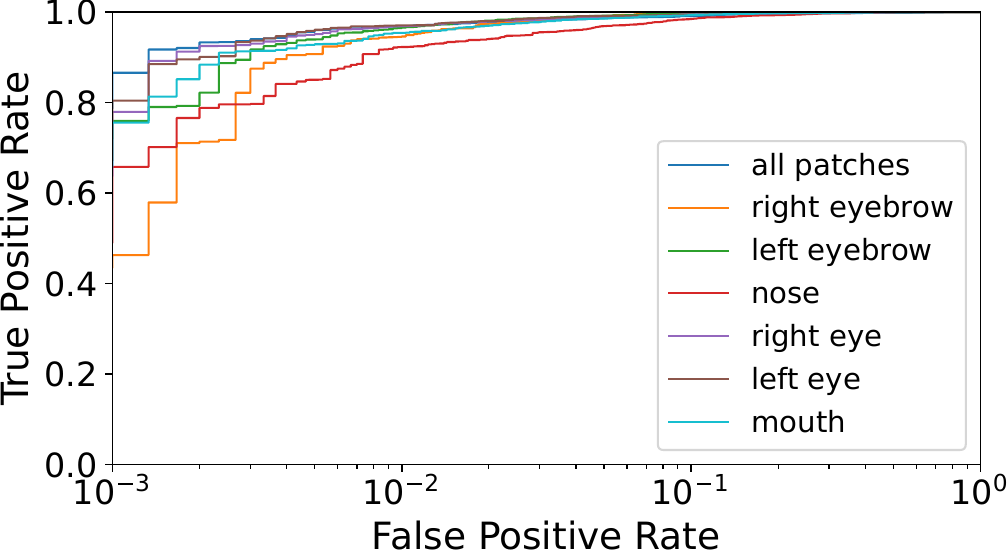}
	\includegraphics[width=0.325\linewidth]{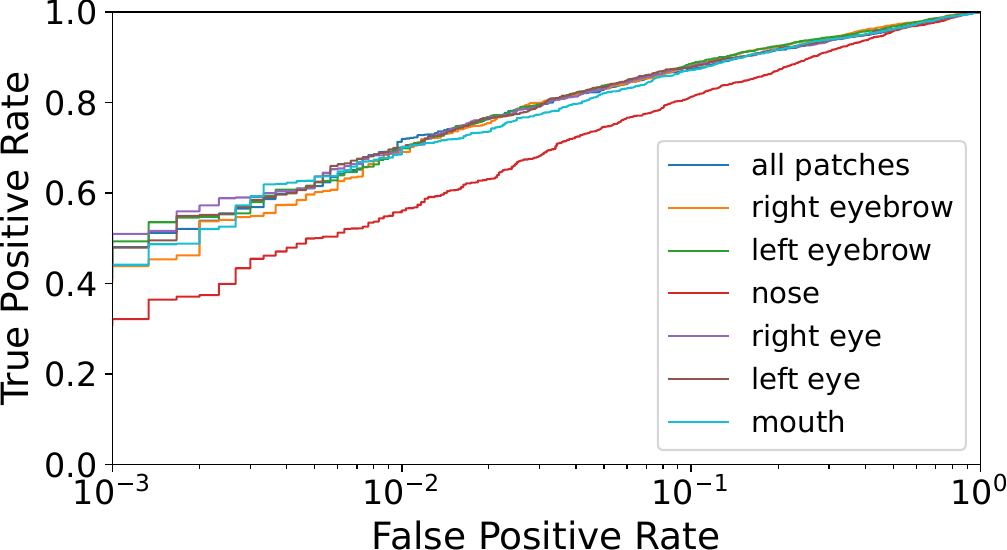}
	\includegraphics[width=0.325\linewidth]{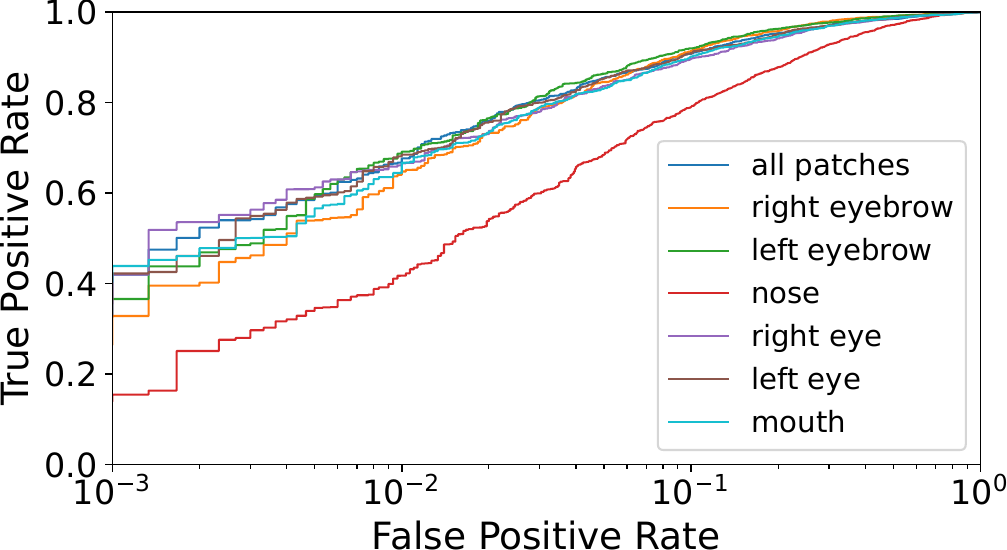}
	
	\textit{(d) LFW} \hspace{120pt}   \textit{(f) CALFW} \hspace{120pt}   \textit{(e) AgeDB}
	\caption{ROC curves of VOIDFace Detach-Patch models (V1-a, b, c; V2-d, f, e) on various benchmarks.}
	\label{fig:performance_12}
\end{figure*}

\section*{Additional Experimentation and Discussion}

\subsection*{Information content within the shares}

Even though VOIDFace shares are visually protected (see Figure 2 in the main paper), evaluating the information content within these shares is crucial. We have used entropy as a metric to assess the information content within the shares. Entropy quantifies the amount of uncertainty or randomness in an image, measuring the diversity of pixel values and their distribution. Higher entropy indicates high randomness, greater complexity and less information content. In this experiment, we assess the richness of an image, which is essential for analyzing the information within the VOIDFace shares. The maximum possible entropy value for 8-bit (uint8) data is eight. Table \ref{tab:entropy} shows the average entropy calculated for the selected VGGFace2 (FRR = 0.05) dataset across different channels. These high values indicate that the shares exhibit significant randomness, revealing no useful information about the corresponding patch.

\begin{table}[!htb]
	\centering
	\scriptsize
	\caption{Average entropy of the shares from different patches.}
	\label{tab:entropy}
	\begin{tabular}{lcccc}
		\hline
		\multirow{2}{*}{Patch} & \multicolumn{3}{c}{Color channels} & \multirow{2}{*}{Average} \\ \cline{2-4}
		& R          & G         & B         &                          \\ \hline
		Left eyebrow           & 7.9802     & 7.9812    & 7.9828    & 7.9814                   \\
		Right eyebrow          & 7.9838     & 7.9838    & 7.9849    & 7.9841                   \\
		Left eye               & 7.9837     & 7.9823    & 7.9805    & 7.9821                   \\
		Right eye              & 7.9791     & 7.9829    & 7.9817    & 7.9812                   \\
		Nose                   & 7.9837     & 7.9825    & 7.9817    & 7.9826                   \\
		Mouth                  & 7.9796     & 7.979     & 7.9809    & 7.9799                   \\ \hline
	\end{tabular}
\end{table}

\subsection*{Correlation coefficient}

In image processing, correlation coefficient plays a key role in evaluating encryption strength. A good image encryption algorithm should produce cipher images with a near-zero correlation between adjacent pixels (horizontally, vertically, and diagonally). The correlation close to zero demonstrates the difficulty of predicting adjacent pixels. In VOIDFace, as the cipher images are shares, we have evaluated the correlation coefficient between the patches and corresponding generated shares in different direction. 
The values close to zero in Table \ref{tab:corr} emphasize the encryption strength and near-impossibility of predicting adjacent pixel values in any direction. Given the large sample space and low correlation coefficients, a brute-force attack on VOIDFace is virtually impossible.

\begin{table}[!htp]
	\centering
	\scriptsize
	\caption{The average correlation coefficients of the shares.}
	\label{tab:corr}
	\begin{tabular}{lccc}
		\hline
		& Horizontal & Vertical &Diagonal \\ \hline
		Left eyebrow & -0.006     & -0.016   & 0.029   \\ 
		Right eyebrow & 0.007     & 0.015   & -0.027   \\ 
		Left eye & -0.017     & -0.009   & -0.031  \\ 
		Right eye & -0.004     & 0.012   & -0.009  \\ 
		Nose & 0.009      & 0.027    & -0.029  \\ 
		Mouth & -0.031     & 0.018    & -0.026  \\ \hline
	\end{tabular}
\end{table}

\subsection*{Patch exclusion analysis - Single Patches}

In further experiment, we evaluated the performance of individual facial patches within the VOIDFace framework. To assess the contribution of a single patch, all other patches were left blank prior to inference by the VOIDFace network (see Figure \ref{fig:performance_11}).

Several key observations emerged from our results. Across all configurations, the "nose" patch consistently achieved the highest accuracy across benchmarks, highlighting the discriminative strength of this region. This aligns with existing understanding that the nose, being one of the least deformable facial features, retains stable and informative characteristics. For images captured in unconstrained conditions (e.g., the LFW benchmark), patches corresponding to the eyebrows and mouth also proved highly informative. Notably, under patch supervision (V2) in the CALFW and AgeDB benchmarks, the mouth patch outperformed those of the eyebrows and eyes, indicating its increased discriminative value in these settings. In other configurations, the various facial regions demonstrated relatively comparable performance.

\subsection*{Patch exclusion analysis - Patches Detach}

In an additional experiment, we simulated the removal of a single facial patch from the input to assess its individual contribution to overall performance (see Figure \ref{fig:performance_12}). Specifically, we repeated the evaluation protocols with one patch replaced by a blank input. The results of this experiment align closely with the findings from the single-patch evaluation described above, further confirming the strong discriminative power of the "nose" region. Notably, excluding the "nose" patch resulted in a significant drop in performance. In contrast, the exclusion of either eye region typically did not reduce performance, while the removal of the mouth or eyebrow patches led to only a minor performance decline.


\end{document}